# Principal Component Analysis-Linear Discriminant Analysis Feature Extractor for Pattern Recognition

Aamir Khan[1], Hasan Farooq[2]

[1] Electrical Engineering Department, COMSATS Institute of Information Technology,
Wah Cantt., Punjab, Pakistan

[2] Electrical Engineering Department, COMSATS Institute of Information Technology,
Wah Cantt., Punjab, Pakistan

**Abstract**

Robustness of embedded biometric systems is of prime importance with the emergence of fourth generation communication devices and advancement in security systems This paper presents the realization of such technologies which demands reliable and error-free biometric identity verification systems. High dimensional patterns are not permitted due to eigen-decomposition in high dimensional image space and degeneration of scattering matrices in small size sample. Generalization, dimensionality reduction and maximizing the margins are controlled by minimizing weight vectors. Results show good pattern by multimodal biometric system proposed in this paper. This paper is aimed at investigating a biometric identity system using Principal Component Analysis and Lindear Discriminant Analysis with K-Nearest Neighbor and implementing such system in real-time using SignalWAVE.

## 1. Introduction

Visual client recognition system is one of the multimodal biometric systems. The system automatically recognizes or identifies the user based on facial information. First, a client base must be enrolled in the system so that a biometric template or reference can be captured. This is used for matching when an individual needs to be identified. Depending on the context, visual recognition system can operate either in verification (authentication) or an identification mode.

The components that are used to design this system are:

1.) Principle Component Analysis (PCA)
2.) Linear Discriminant Analysis (LDA)
3.) Nearest Neighbor Classifier(NN)

After 100% successful simulation results of the multimodal system proposed, same system was implemented onto the Field Programmable Gate Array (FPGA) using SignalWave.

**PCA:** Principal component analysis is a statistical tool used to analyze data sets. The central idea of principal component analysis (PCA) is to reduce the dimensionality of a data set consisting of large number of interrelated variables, while retaining as much as possible of the variation present in the data set [1]. The mathematics behind principle component analysis is statistics and is hinged behind standard deviation, eigenvalues and eigenvectors. The entire subject of statistics is based around the idea that you have this big set of data, and you want to analyze that set in terms of the relationships between the individual points in that data set [2]. Images are technically data set whose component represents the image which we see.

**LDA:** The standard LDA can be seriously degraded if there are only a limited number of observations $N$ compared to the dimension of the feature space $n$ [5]. To prevent this from happen is it is recommended that the linear discriminant analysis be preceded by a principle component analysis. In PCA, the shape and location of the original data sets changes when transformed to a different space whereas LDA doesn't change the location but only tries to provide more class separability and draw a decision region between the given classes [6].

**k-Nearest Neighbor:** The projected training vectors are subtracted from the projected test sound vector and squared to get the distance. The sum of each column and its square roots, mean and minimum value in array is calculated.

## 2. Methodology

Each 2-D facial image is represented as 1-D vector by concatenating each column (or row) into a long thin vector. So, the resulting vector should present as:

$$x^i = [x_1^i \ldots x_N^i]^T$$

The mean image is a column vector where each entry is the mean of all corresponding pixels of the training images. The preceding theory can be expressed in the following expression:

$$\bar{x}^i = x^i - m$$

Where

$$m = \frac{1}{p}\sum_{i=1}^{p} x^i$$

Center data

$$\bar{X} = \lfloor \bar{x}^1 | \bar{x}^2 | \ldots | \bar{x}^p | \rfloor$$

Covariance matrix

$$\Omega = \overline{XX}^T$$

Sorting the order of Eigenvectors

$$V = \lfloor V_1 | V_2 | \ldots | V_p | \rfloor$$

Projection of the image

$$\tilde{x}^i = V^T \bar{x}^i$$

Identifying the new images

$$\bar{y}^i = y^i - m$$

and

$$\tilde{y}^i = V^T \bar{y}^i$$

After using PCA, unwanted variations caused by the illumination, facial position and facial expression still retain. Accordingly, the features produced by PCA are not necessarily good for discriminant among classes. Therefore, the most discriminating face features are further acquired by using the LDA by calculating the Class Matrix, Between Class Matrix, Generalized Eigenvalue and Eigenvector, sorting the order of eigenvector and projecting training images onto the fisher basis vectors.

$$S_i = \sum_{x \in X_i}(x - m_i)(x - m_i)^T$$

$$S_W = \sum_{i=1}^{C} S_i$$

Where C is the number of classes

$$S_B = \sum_{i=1}^{C} n_i (x - m_i)(x - m_i)^T$$

Where $n_i$ is the number of image in the *ith* class, *m* is the total mean of all the training images.

Euclidean distance is used to compute the distance. The mathematic formula of Euclidean distance is presented as:

$$D(x_j, y_j) = \sqrt{\sum_{j=1}^{n}(x_j - y_j)^2}$$

This equation would generate the Euclidean distant matrix. Euclidean distance matrix, is a table of distance-square between points (in this case $x_j$ and $y_j$) taken by pair from a list of a defined number of points.

By comparing the output of this operation to a set of sample result we identified five clients in the data set.

For implementation of the system on FPGA using SignalWave, most of the blocks were taken from the workspace especially in the PCA and LDA Simulink blockset. The Nearest neighbor simulink block was designed entirely from the program. Complete visual recognition system is shown in figure 7 in appendix. As Simulink is an integral part of MATLAB, it is easy to switch back and forth during the design though the logical interconnects from the program is of paramount importance. When performing real-time operation using this design all the results obtained were correct.

## 3. Results

By PCA and LDA, projected new image is then used to compare with each of the training image projection. The training image that is closest to the new image is used to identify the new image. This is represented in a plot (figure 1 and figure 2).

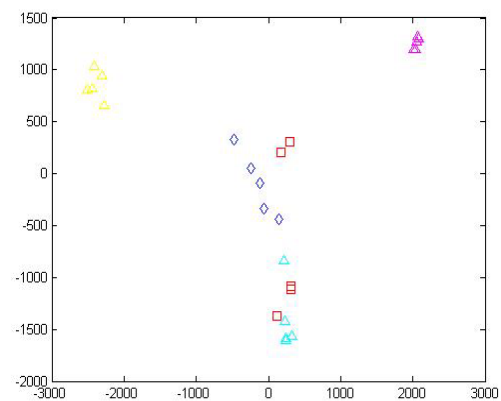

Fig. 1 PCA plot showing 5 clients in no particular order.

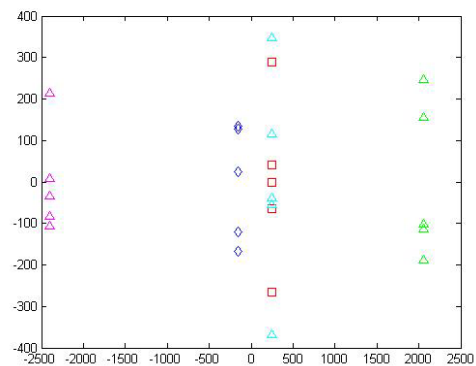

Fig. 2    LDA plot.

After making each block in Simulink, constant referral was made to the workpsace in order to match the output from simulink blocks to the output from the code. The final result which was obtained at the display block is shown below.

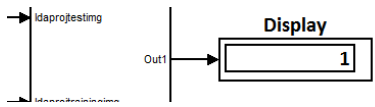

Fig. 3 Result for k=2 (test image database)

For K=5:

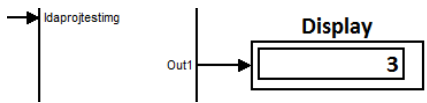

Fig. 4 Result for k=5 (test image database)

For K=10:

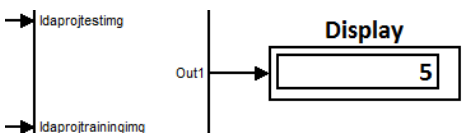

Fig. 5 Result for k=10 (test image database)

In obtaining image from the fraud database and K=3, therefore the following result is obtained:

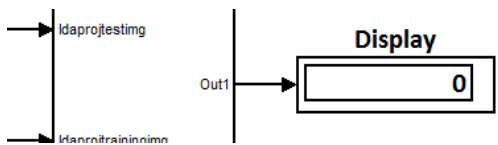

Fig. 6 Result for k=2 (fraud database)

## 4. Conclusion

The multimodal biometric system presented in this paper using PCA and LDA along-with k-Nearest Neighbor (kNN). The paper also delineates a feasible solution for implementing the proposed system on FPGA for significant speed increase. The efficiency highly increases with the use of LDA over PCA. Visual recognition system proposed in this paper based on real tests give accurate results and significantly decreases the complexity. It is observed during the whole study that if PCA and LDA are both used jointly it produce accurate results. Furthermore, it is recommended to experiment LDA with support vector machine (SVM) for speech recognition.

**Appendix**

System Design

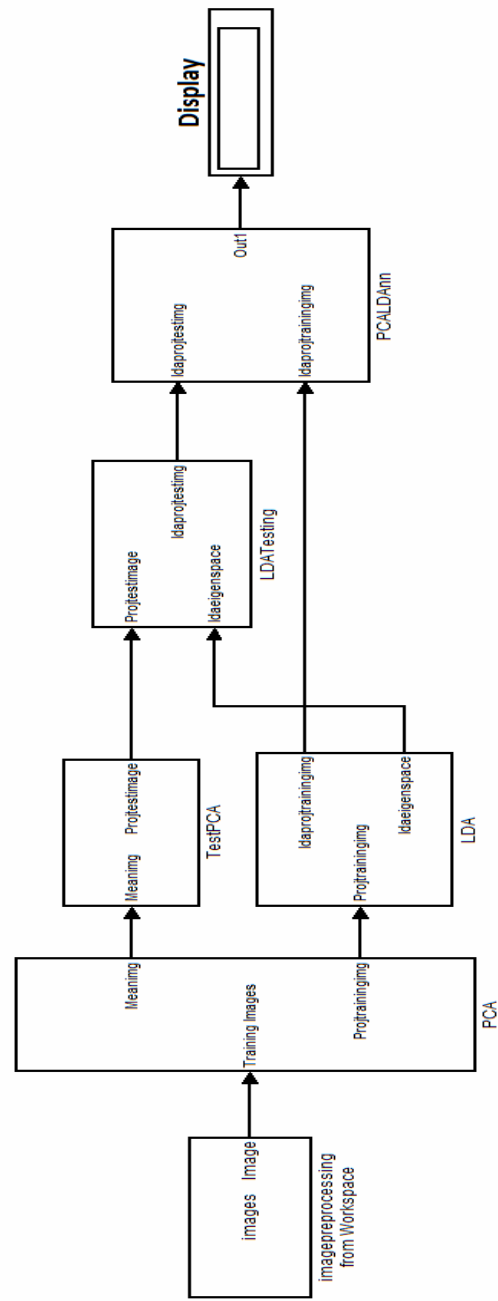

**Aamir Khan** has received M.Sc. degree in Electronic Communications and Computer Engineering from University of Nottingham Malaysia Campus in 2011. He is working at CIIT Wah Campus as lecturer in Electrical Engineering Department. Research interests include embedded systems, intelligent systems and optical communications.

**Hasan Farooq** has received his degree BSc. Degree in Electrical Engineering from University of Engineering and Technology Lahore Pakistan in 2009.s Electrical Engineering Department. He is working as research assistant at CIIT Wah Campus in it Research interests include optical communications, sensor and cognitive networks.